\documentclass[runningheads]{llncs}
\usepackage{graphicx}
\usepackage{amsmath,amssymb} % define this before the line numbering.
\usepackage{color}
\usepackage{multirow}
\usepackage{url}
\usepackage{color}
\usepackage{caption}
\usepackage{booktabs}
\usepackage{multirow}
\usepackage{array}
\usepackage{lettrine}
\usepackage{mdwlist}
\usepackage{multirow}
\usepackage{bbding}
\usepackage{caption}
\usepackage{xspace}
\usepackage{paralist}

\usepackage[width=122mm,left=12mm,paperwidth=146mm,height=193mm,top=12mm,paperheight=217mm]{geometry}

\def\comment#1{{}}
\def\eg{{\em e.g.}}
\def\ie{{\em i.e.}}
\def\etal{{\em et al.}}
\def\uav{UAVDT\xspace} % name of dataset
\begin{document}
% \linenumbers
\pagestyle{headings}
\mainmatter

\title{The Unmanned Aerial Vehicle Benchmark:\\ Object Detection and Tracking} % Replace with your title

\titlerunning{arXiv}

\authorrunning{arXiv}

\author{$^1$Dawei Du, $^2$Yuankai Qi, $^2$Hongyang Yu, $^1$Yifan Yang, $^1$Kaiwen Duan, $^1$Guorong Li, $^3$Weigang Zhang, $^1$Qingming Huang, $^4$Qi Tian}

%Please write out author names in full in the paper, i.e. full given and family names.
%If any authors have names that can be parsed into FirstName LastName in multiple ways, please include the correct parsing, in a comment to the volume editors:
%\index{Lastnames, Firstnames}
%(Do not uncomment it, because you may introduce extra index items if you do that...)

\institute{$^1$University of Chinese Academy of Sciences\\
	$^2$Harbin Institute of Technology\\
    $^3$Harbin Institute of Technology, Weihai\\
    $^4$The University of Texas at San Antonio\\
	\email{dawei.du@vipl.ict.ac.cn}
}

\maketitle

\begin{abstract}
  With the advantage of high mobility, Unmanned Aerial Vehicles (UAVs) are used to fuel numerous important applications in computer vision, delivering more efficiency and convenience than surveillance cameras with fixed camera angle, scale and view. However, very limited UAV datasets are proposed, and they focus only on a specific task such as visual tracking or object detection in relatively constrained scenarios. Consequently, it is of great importance to develop an unconstrained UAV benchmark to boost related researches. In this paper, we construct a new UAV benchmark focusing on complex scenarios with new level challenges. Selected from $10$ hours raw videos, about $80,000$ representative frames are fully annotated with bounding boxes as well as up to $14$ kinds of attributes (\eg, weather condition, flying altitude, camera view, vehicle category, and occlusion) for three fundamental computer vision tasks: object detection, single object tracking, and multiple object tracking. Then, a detailed quantitative study is performed using most recent state-of-the-art algorithms for each task. Experimental results show that the current state-of-the-art methods perform relative worse on our dataset, due to the new challenges appeared in UAV based real scenes, \eg, high density, small object, and camera motion. To our knowledge, our work is the first time to explore such issues in unconstrained scenes comprehensively.
\end{abstract}

\section{Introduction}

With the fast development of artificial intelligence, higher request to efficient and effective intelligent vision systems is putting forward. To tackle with higher semantical tasks in computer vision, such as object recognition, behavior analysis and motion analysis, researchers have developed numerous fundamental detection and tracking algorithms for the past decades.
%-------------------------------------------------------------------------
\begin{table*}[ht]
\centering
      \setlength{\tabcolsep}{0.5pt}
      \scriptsize{
    \begin{tabular}{c|c|c|c|c|c|c|c|c|c|c}
    \hline
    \multirow{2}{*}{Datasets} &\multicolumn{9}{c}{Attributes} \\
    \cline{2-11} &UAV &Frames &Boxes &Tasks &Vehicles &Weather &Occlusion &Altitude &View &Year \\
    \hline
    MIT-Car~\cite{DBLP:journals/ijcv/PapageorgiouP00}              & &$1.1k$ &$1.1k$ &D &\Checkmark &  & & & &2000 \\
    Caltech~\cite{DBLP:journals/pami/DollarWSP12}                  & &$132k$ &$347k$ &D & & &\Checkmark  & & &2012 \\
    KAIST~\cite{DBLP:conf/cvpr/HwangPKCK15}                        & &$95k$ &$86k$   &D & &\Checkmark &\Checkmark & & &2015 \\
    KITTI-D~\cite{DBLP:conf/cvpr/GeigerLU12}                       & &$15k$ &$80.3k$ &D &\Checkmark & &\Checkmark & & &2014 \\
    MOT17Det~\cite{MOT17}                                          & &$11.2k$ &$392.8k$ &D & & &\Checkmark & & &2017 \\
    CARPK~\cite{DBLP:conf/iccv/HsiehLH17}                          &\Checkmark &$1.5k$ &$90k$ &D &\Checkmark & & & & &2017 \\
    Okutama~\cite{DBLP:conf/cvpr/BarekatainMSMNM17}         &\Checkmark &$77.4k$ &$422.1k$ &D & & & & & &2017 \\
    PETS2009~\cite{DBLP:conf/avss/FerrymanE09}                     &  &$1.5k$ &$18.5k$ &D,M & &\Checkmark & & & &2009 \\
    KITTI-T~\cite{DBLP:conf/cvpr/GeigerLU12}                       & &$19k$ &$>47.3k$  &M &\Checkmark & &\Checkmark & & &2014 \\
    MOT15~\cite{DBLP:journals/corr/Leal-TaixeMRRS15}               & &$11.3k$ &$>101k$ &M & &\Checkmark  & & & &2015 \\
    DukeMTMC~\cite{DBLP:conf/eccv/RistaniSZCT16}                   & &$2852.2k$ &$4077.1k$ &M & &  &\Checkmark & & &2016 \\
    DETRAC~\cite{DBLP:journals/corr/WenDCLCQLYL15}                 & &$140k$ &$1210k$  &D,M &\Checkmark &\Checkmark  &\Checkmark  & & &2016 \\
    Campus~\cite{DBLP:conf/eccv/RobicquetSAS16}                    &\Checkmark &$929.5k$ &$19.5k$ &M &\Checkmark & & & & &2016 \\
    MOT16~\cite{DBLP:journals/corr/Anton16}                        & &$11.2k$ &$>292k$ &M & &\Checkmark  &\Checkmark & & &2016 \\
    MOT17~\cite{MOT17}                                             & &$11.2k$ &$392.8k$ &M & &\Checkmark  &\Checkmark & & &2017 \\
    ALOV300~\cite{DBLP:journals/pami/SmeuldersCCCDS14}             & &$151.6k$ &$151.6k$ &S & & & & & &2015 \\
    OTB100~\cite{DBLP:journals/pami/WuLY15}                        & &$59k$ &$59k$ &S & & & & & &2015 \\
    VOT2016~\cite{DBLP:conf/eccv/KristanLMFPCVHL16}                & &$21.5k$ &$21.5k$ &S & & &\Checkmark  & & &2016 \\
    UAV123~\cite{DBLP:conf/eccv/MuellerSG16}                       &\Checkmark &$110k$ &$110k$ &S &\Checkmark & & & & &2016 \\
  \hline
  \uav    &\Checkmark &$80k$ &$841.5k$ &D,M,S &\Checkmark &\Checkmark &\Checkmark &\Checkmark &\Checkmark &2017 \\
  \hline
\end{tabular}}
\caption{Summary of existing related datasets ($1k=10^{3}$). D=DET, M=MOT, S=SOT.}
\label{tab:comparison-dataset}
\end{table*}

To evaluate these algorithms fairly, the computer vision community has developed plenty of datasets including detection datasets (\eg, Caltech~\cite{DBLP:journals/pami/DollarWSP12} and DETRAC~\cite{DBLP:journals/corr/WenDCLCQLYL15}) and tracking datasets (\eg, KITTI-T~\cite{DBLP:conf/cvpr/GeigerLU12} and VOT2016~\cite{DBLP:conf/eccv/KristanLMFPCVHL16}). The common shortcoming of these datasets is that videos are captured by fixed or moving car based cameras, which is limited in viewing angles in surveillance scene.

Benefiting from flourishing global drone industry, Unmanned Aerial Vehicle (UAV) has been applied in many areas such as security and surveillance, search and rescue, and sports analysis. Different from traditional surveillance cameras, UAV with moving camera has several advantages inherently, such as easy to deploy, high mobility, large view scope, and uniform scale. Thus it brings new challenges to existing detection and tracking technologies, such as:
\begin{itemize}
  \item \textbf{High Density.} Since UAV cameras are flexible to capture videos at wider view angle than fixed cameras, leading to large object number.
  \item \textbf{Small Object.} Objects are usually small or tiny due to high altitude of UAV views, resulting in difficulties to detect and track them.
  \item \textbf{Camera Motion.} Objects move very fast or rotate drastically due to the high-speed flying or camera rotation of UAVs.
  \item \textbf{Realtime Issues.} The algorithms should consider realtime issues and maintain comparable accuracy on embedded UAV platforms for practical application.
\end{itemize}
To study these problems, limited UAV datasets are collected such as Campus~\cite{DBLP:conf/eccv/RobicquetSAS16} and CARPK~\cite{DBLP:conf/iccv/HsiehLH17}. However, they only focus on a specific task such as visual tracking or detection in constrained scenes, for instance campus or parking lots. The community needs a more comprehensive UAV benchmark in unconstrained scenarios for further boosting research on related tasks.

To this end, we construct a large scale challenging UAV Detection and Tracking (\uav) benchmark (\ie, about $80,000$ representative frames from $10$ hours raw videos) for $3$ important fundamental tasks, \ie, object DETection (DET), Single Object Tracking (SOT) and Multiple Object Tracking (MOT). Our dataset is captured by UAVs\footnote{We use DJI Inspire 2 to collect videos, and more information about the UAV platform can be found in~\url{http://www.dji.com/inspire-2}.} in various complex scenarios. Since the current majority of datasets focus on pedestrians, as a supplement, the objects of interest in our benchmark are \textit{vehicles}. Moreover, these frames are manually annotated with bounding boxes and some useful attributes, \eg, vehicle category and occlusion. This paper makes the following contributions: (1) We collect a fully annotated dataset for $3$ fundamental tasks applied in UAV surveillance. (2) We provide an extensive evaluation of the most recently state-of-the-art algorithms in various attributes for each task.
\section{\uav Benchmark}
The \uav benchmark\footnote{We will release the dataset and all the experimental results upon the acceptance of our paper.} consists of $100$ video sequences, which are selected from over $10$ hours of videos taken with an UAV platform at a number of locations in urban areas, representing various common scenes including squares, arterial streets, toll stations, highways, crossings and T-junctions. The videos are recorded at $30$ frames per seconds (fps), with the JPEG image resolution of $1080\times540$ pixels.
\subsection{Data Annotation}
\begin{figure*}[t]
\centering
\includegraphics[width=.95\linewidth]{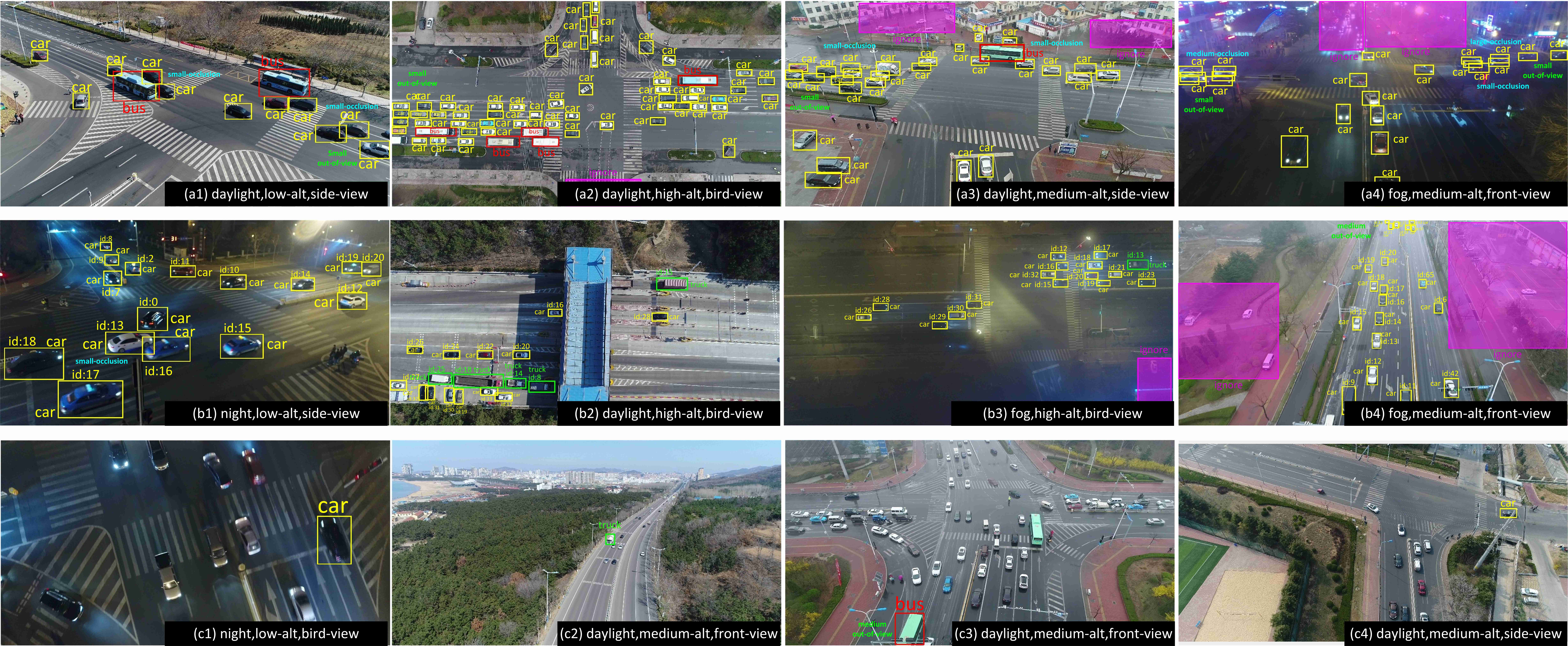}
  \caption{Examples of annotated frames in the \uav benchmark. The three rows indicate the DET, MOT and SOT task, respectively. The shooting conditions of UAVs are presented in the lower right corner. The pink areas are ignored regions in the dataset. Different bounding box colors denote different classes of vehicles. For clarity, we only display some attributes.}
\label{fig_anno}
\end{figure*}

For annotation, we ask over $10$ domain experts to label our dataset using the vatic tool\footnote{\url{http://carlvondrick.com/vatic/}} for two months. With several rounds of double-check, the annotation errors are reduced as much as possible. Specifically, about $80,000$ frames in the \uav benchmark dataset are annotated over $2,700$ vehicles with $0.84$ million bounding boxes. According to PASCAL VOC~\cite{DBLP:journals/ijcv/EveringhamEGWWZ15}, the regions that cover too small vehicles are ignored in each frame due to low resolution. Figure~\ref{fig_anno} shows some sample frames with annotated attributes in the dataset.

Based on different shooting conditions of UAVs, we first define $3$ attributes for MOT task:
\begin{itemize}
  \item \textbf{Weather Condition} indicates illumination when capturing videos, which affects appearance representation of objects. It includes \textit{daylight}, \textit{night} and \textit{fog}. Specifically, videos shot in daylight introduce interference of shadows. Night scene, bearing dim street lamp light, offers scarcely any texture information. In the meantime, frames captured at \textit{fog} lack sharp details so that contours of objects vanish in the background.

  \item \textbf{Flying Altitude} is the flying height of UAVs, affecting the scale variation of objects. Three levels are annotated, \ie, \textit{low-alt}, \textit{medium-alt} and \textit{high-alt}. When shooting in low-altitude ($10m\sim30m$), more details of objects are captured. Meanwhile the object may occupy larger area, \eg, $22.6\%$ pixels of a frame in an extreme situation. When videos are collected in medium-altitude ($30m\sim70m$), more view angles are presented. While in much higher altitude ($>70m$), plentiful vehicles are of less clarity. For example, most tiny objects just contain $0.005\%$ pixels of a frame, yet object numbers can be more than a hundred.

  \item \textbf{Camera View} consists of $3$ object views. Specifically, \textit{front-view}, \textit{side-view} and \textit{bird-view} mean the camera shooting along with the road, on the side, on the top of objects, respectively. Note that the first two views may coexist in one sequence.
\end{itemize}

\begin{figure}[t]
\centering
\includegraphics[width=.75\linewidth]{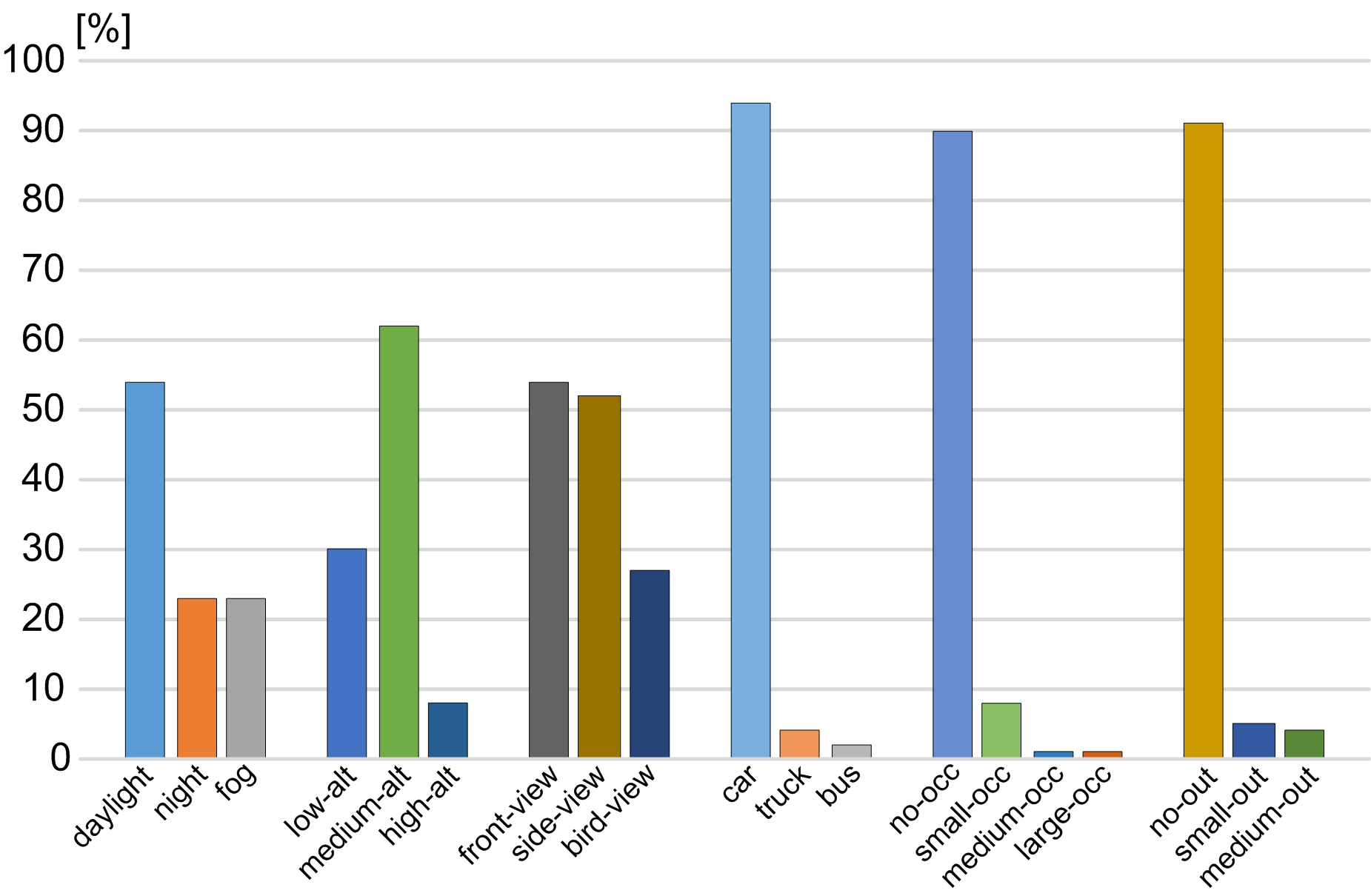}
  \caption{The distribution of attributes of both DET and MOT tasks in \uav.}
\label{fig_attr_overall}
\end{figure}

To evaluate DET algorithms thoroughly, we also label another $3$ attributes including \textit{vehicle category}, \textit{vehicle occlusion} and \textit{out-of-view}. vehicle category consists of \textit{car}, \textit{truck} and \textit{bus}. vehicle occlusion is the fraction of bounding box occlusion, \ie, \textit{no-occ} ($0\%$), \textit{small-occ} ($1\%\sim30\%$), \textit{medium-occ} ($30\%\sim70\%$) and \textit{large-occ} ($70\%\sim100\%$). Out-of-view indicates the degree of vehicle parts outside frame, divided into \textit{no-out} ($0\%$), \textit{small-out} ($1\%\sim30\%$) and \textit{medium-out} ($30\%\sim50\%$). The objects are discarded when the out-of-view ratio is larger than $50\%$. The distribution of the above attributes is shown in Figure~\ref{fig_attr_overall}. Within an image, objects are defined as ``occluded'' by other objects or the obstacles in the scenes, \eg, under the bridge; while objects are regarded as ``out-of-view'' when they are out of the image or in the ignored regions.

For SOT task, $8$ attributes are annotated for each sequence, \ie, Background Clutter (\textbf{BC}), Camera Rotation (\textbf{CR}), Object Rotation (\textbf{OR}), Small Object (\textbf{SO}), Illumination Variation (\textbf{IV}), Object Blur (\textbf{OB}), Scale Variation (\textbf{SV}) and Large Occlusion (\textbf{LO}). The distribution of SOT attributes is presented in Table~\ref{tab:sot_attr_sta}. Specifically, $74\%$ videos contain at least $4$ visual challenges, and among them $51\%$ have $5$ challenges. Meanwhile, $27\%$ of frames contribute to long-term tracking videos. As a consequence, a candidate SOT method can be estimated in various cruel environment, most likely at the same frame, guaranteeing the objectivity and discrimination of the proposed dataset.
\begin{table}
\centering
\footnotesize\setlength{\tabcolsep}{5pt}
\begin{tabular}{|c|c|c|c|c|c|c|c|c|}
  \hline
  &\textbf{BC} &\textbf{CR} &\textbf{OR} &\textbf{SO} &\textbf{IV} &\textbf{OB} &\textbf{SV} &\textbf{LO}\\\hline
  \textbf{BC}  &$\textbf{29}$	&$18$	&$20$	&$12$	&$17$	&$9$	&$16$	&$18$\\
  \textbf{CR}  &$18$	&$\textbf{30}$	&$21$	&$14$	&$17$	&$12$	&$18$	&$12$\\
  \textbf{OR}  &$20$	&$21$	&$\textbf{32}$	&$12$	&$17$	&$13$	&$23$	&$14$\\
  \textbf{SO}  &$12$	&$14$	&$12$	&$\textbf{23}$	&$13$	&$13$	&$8$	&$6$\\
  \textbf{IV}  &$17$	&$17$	&$17$	&$13$	&$\textbf{28}$	&$18$	&$12$	&$7$\\
  \textbf{OB}  &$9$	&$12$	&$13$	&$13$	&$18$	&$\textbf{23}$	&$11$	&$2$\\
  \textbf{SV}  &$16$	&$18$	&$23$	&$8$	&$12$	&$11$	&$\textbf{29}$	&$14$\\
  \textbf{LO}  &$18$	&$12$	&$14$	&$6$	&$7$	&$2$	&$14$	&$\textbf{20}$\\
  \hline
\end{tabular}
\caption{Distribution of SOT attributes, showing the number of coincident attributes across all videos. The diagonal line denotes the number of sequences with only one attribute.}
\label{tab:sot_attr_sta}
\end{table}

Notably, our benchmark is divided into training and testing sets, with $30$ and $70$ sequences, respectively. The testing set consists of $20$ sequences for both DET and MOT tasks, and $50$ for SOT task. Besides, training videos are taken at different locations from the testing videos, but share similar scenes and attributes. This setting reduces the overfitting probability to particular scenario.
\subsection{Comparison with Existing UAV Datasets}
Although new challenges are brought to computer vision by UAVs, limited datasets~\cite{DBLP:conf/eccv/MuellerSG16,DBLP:conf/eccv/RobicquetSAS16,DBLP:conf/iccv/HsiehLH17} have been published to accelerate the improvement and evaluation of various vision tasks. By exploring the flexibility of UAV¡¯s flare maneuver in both altitude and plane domain, Matthias~\etal~\cite{DBLP:conf/eccv/MuellerSG16} propose a low-altitude UAV tracking dataset to evaluate ability of SOT methods of tackling with relatively fierce camera movement, scale change and illumination variation, yet it still lacks varieties in both weather conditions and camera motions, and its scenes are much less clustered than real circumstances. In~\cite{DBLP:conf/eccv/RobicquetSAS16}, several video fragments are collected to analyze the behaviors of pedestrians in top-view scenes of campus with fixed UAV cameras for the MOT task. Although ideal visual angles benefit trackers to obtain stable trajectories by narrowing down challenges they have to meet, it also risks diversity while evaluating MOT methods. Hsieh~\etal~\cite{DBLP:conf/iccv/HsiehLH17} present an annotated dataset aiming at counting vehicles in parking lots. However, our dataset captures videos in unconstrained areas, resulting in more generalization.

The detail comparisons of the proposed dataset with other related works are summarized in Table~\ref{tab:comparison-dataset}. Although the proposed dataset is not the largest one compared to existing datasets, it can represent the characteristics of UAV videos more effectively:
\begin{itemize}
  \item Our dataset provides a higher object density $10.52$\footnote{The object density indicates the mean number of objects in each frame.}, compared to related works (\eg, UAV123~\cite{DBLP:conf/eccv/MuellerSG16} $1.00$, Campus~\cite{DBLP:conf/eccv/RobicquetSAS16} $0.02$, DETRAC~\cite{DBLP:journals/corr/WenDCLCQLYL15} $8.64$ and KITTI~\cite{DBLP:conf/cvpr/GeigerLU12} $5.35$). CARPK~\cite{DBLP:conf/iccv/HsiehLH17} is an image based dataset to detect parking vehicles, which is not suitable for visual tracking.
  \item Compared to related works~\cite{DBLP:conf/eccv/MuellerSG16,DBLP:conf/eccv/RobicquetSAS16,DBLP:conf/iccv/HsiehLH17} just focusing on specified scene, our dataset is collected from various scenarios in different weather conditions, flying altitudes, and camera views, \textit{etc}.
\end{itemize}
%-------------------------------------------------------------------------
\section{Evaluation and Analysis}
We run a representative set of state-of-the-art algorithms for each task. Codes for these methods are either available online or from the authors. All the algorithms are trained on the training set and evaluated on the testing set. Interestingly, we find that some high ranking algorithms in other datasets may fail in complex scenarios.
\begin{figure}[t]
\centering
\includegraphics[width=.9\linewidth]{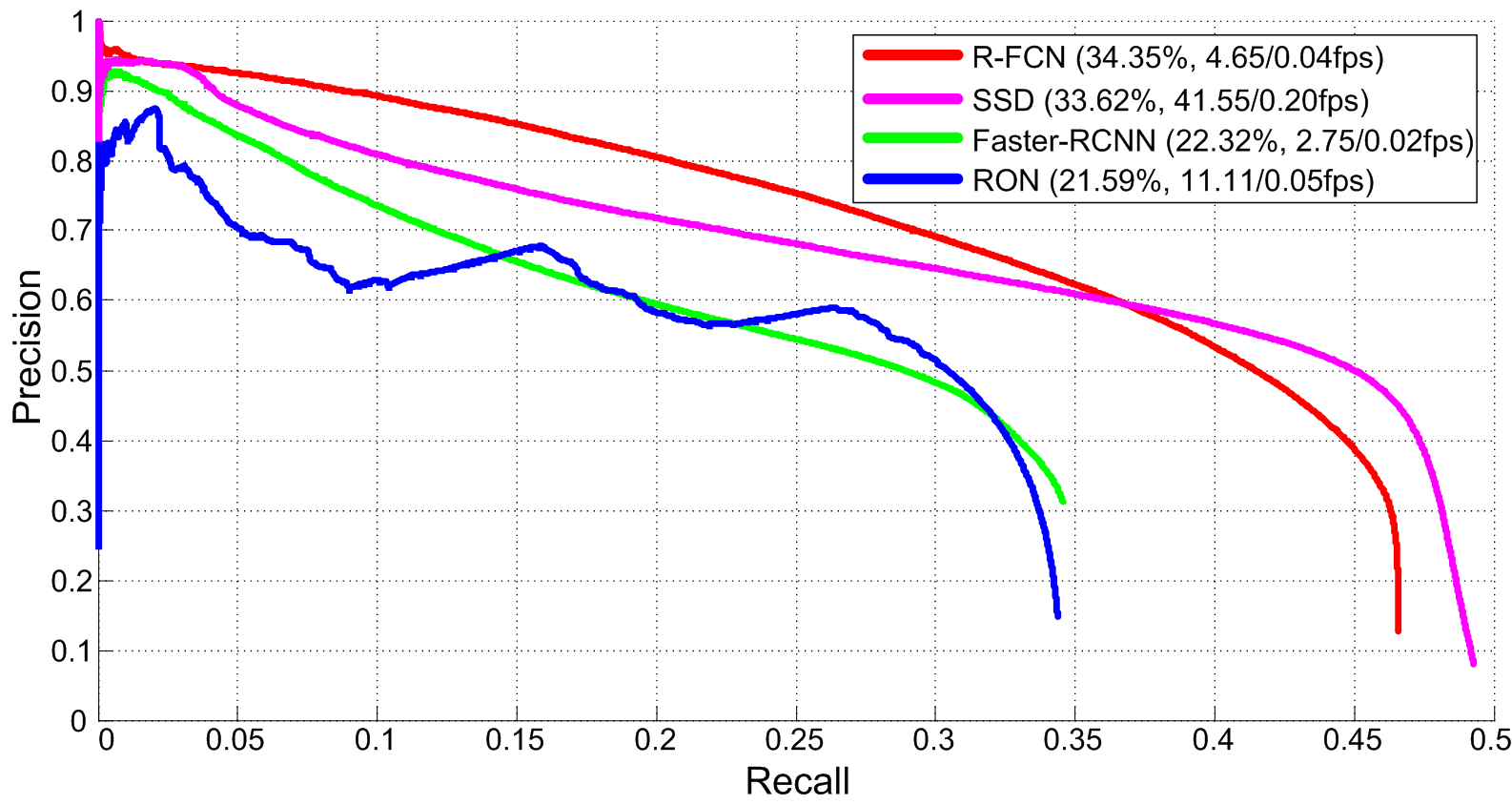}
  \caption{Precision-Recall plot on the testing set of the \uav-DET dataset. The legend presents the AP score and the GPU/CPU speed of each DET method respectively.}
\label{fig_det_overall}
\end{figure}
\subsection{Object Detection}
The current top deep based object detection frameworks is divided into two main categories: region-based (\eg, Faster-RCNN~\cite{DBLP:conf/nips/RenHGS15} and R-FCN~\cite{DBLP:conf/nips/DaiLHS16}) and region-free (\eg, SSD~\cite{DBLP:conf/eccv/LiuAESRFB16} and RON~\cite{DBLP:conf/cvpr/KongSYLLC17}). Therefore, we evaluate the above mentioned $4$ detectors in the \uav dataset.

{\noindent \textbf{Metrics.}} We follow the strategy in the PASCAL VOC challenge~\cite{DBLP:journals/ijcv/EveringhamEGWWZ15} to compute the Average Precision (AP) score in the Precision-Recall plot to rank the performance of DET methods. As performed in KITTI-D~\cite{DBLP:conf/cvpr/GeigerLU12}, the hit/miss threshold of the overlap between a pair of detected and groundtruth bounding boxes is set to $0.7$.

{\noindent \textbf{Implementation Details.}} We train all DET methods on a machine with CPU i9 7900x and 64G memory, as well as a Nvidia GTX 1080 Ti GPU. Faster-RCNN and R-FCN are fine-tuned on the VGG-16 network and Resnet-50, respectively. We use $0.001$ as the learning rate for the first $60k$ iterations and $0.0001$ for the next $20k$ iterations. For region-free methods, the batch size is $5$ for $512\times512$ model according to the GPU capacity. For SSD, we use $0.005$ as the learning rate for $120k$ iterations. For RON, we use the $0.001$ as the learning rate for the first $90k$ iterations, then we decay it to $0.0001$ and continue training for the next $30k$ iterations.  For all the algorithms, we use a momentum of $0.9$ and a weight decay of $0.0005$.
\begin{figure*}[t]
\centering
\includegraphics[width=.95\linewidth]{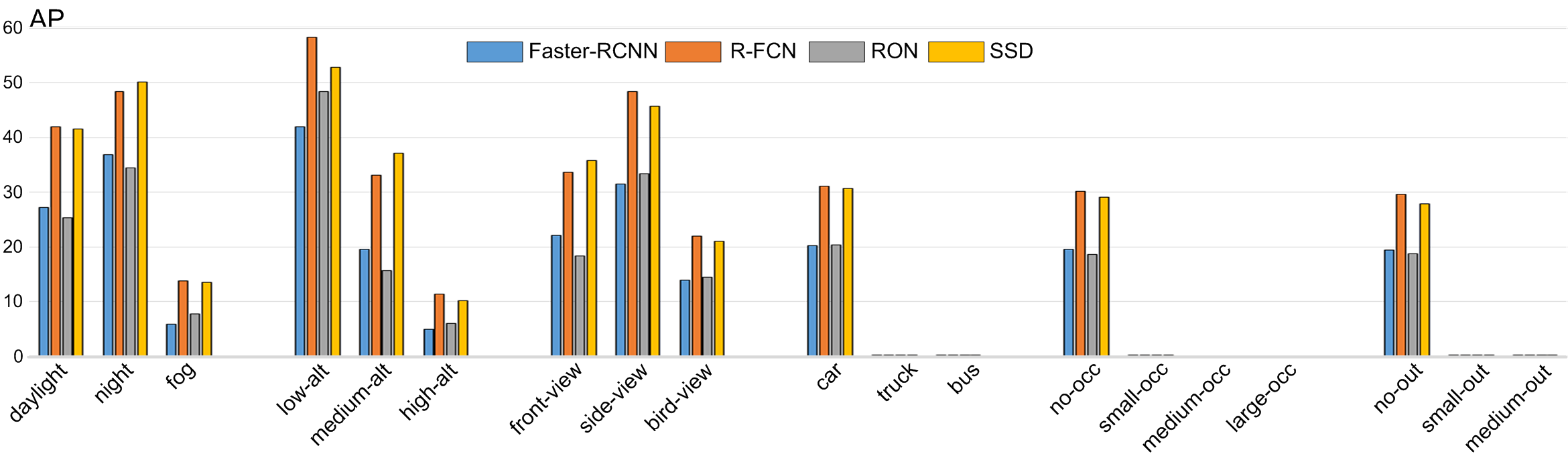}
  \caption{Quantitative comparison results of DET methods in each attribute.}
\label{fig_attr_det}
\end{figure*}
\subsubsection{Overall Evaluation}
Figure~\ref{fig_det_overall} shows the quantitative comparisons of DET methods, which shows no promising accuracy. For example, R-FCN obtains $70.06\%$ AP score even in the hard set of KITTI-D\footnote{The detection result is copied from \url{http://www.cvlibs.net/datasets/kitti/eval_object.php?obj_benchmark=2d}.}, but only $34.35\%$ in our dataset. This maybe our dataset contains a large number of small objects due to the shooting perspective, which is a difficult challenge in object detection. Another reason is that higher altitude brings more cluttered background.

To tackle with this problem, SSD combines multi-scale feature maps to handle objects of various sizes. Yet their feature maps are usually extracted from former layers, which lacks enough semantic meanings for small objects. Improved from SSD, RON fuses more semantic information from latter layers using a reverse connection, and performs well on other datasets such as PASCAL VOC~\cite{DBLP:journals/ijcv/EveringhamEGWWZ15}. Nevertheless, RON is inferior to SSD on our dataset. It maybe because the later layers are so abstract that represent the appearance of small objects not so effectively due to the low resolution. Thus the reverse connection fusing the latter layers may interfere with features in former layers, resulting in inferior performance. On the other hand, region-based methods offer more accurate initial locations for robust results by generating region proposals from region proposal networks. It is worth mentioning that R-FCN achieves the best result by making the unshared per-ROI computation of Faster-RCNN to be sharable~\cite{DBLP:conf/cvpr/KongSYLLC17}.

\subsubsection{Attribute-based Evaluation}
To further explore the effectiveness of DET methods on different situations, we also evaluate them on different attributes in Figure~\ref{fig_attr_det}. For the first $3$ attributes, DET methods perform better on the sequences where objects have more details \eg, \textit{low-alt} and \textit{side-view}. While the object number is bigger and the background is more cluttered in \textit{daylight} than \textit{night}, leading to worse performance in \textit{daylight}. For the remaining attributes, the performance drops very dramatically when detecting large vehicles, as well as handling with occlusion and out-of-view. The results can be attributed to two factors. Firstly, very limited training samples of large vehicles make it hard to train the detector to recognize them. As shown in Figure~\ref{fig_attr_overall}, the number of \textit{truck} and \textit{bus} is only less than $10\%$ of the whole dataset. Besides, it is even harder to detect small objects with other interference. Much work need to be done for small object detection under occlusions or out-of-view.

{\noindent \textbf{Run-time Performance.}} Although region based methods obtain relative good performance, their running speeds (\ie, $<5$fps) are too slow for practical applications especially with constrained computing resources. On the contrary, region free methods save the time of region proposal generation, and proceed at almost realtime speed.
\subsection{Multiple Object Tracking}
MOT methods are generally grouped into online or batch based. Therefore, we evaluate $8$ recent algorithms including online methods (CMOT~\cite{DBLP:conf/cvpr/BaeY14}, MDP~\cite{DBLP:conf/iccv/XiangAS15}, SORT~\cite{DBLP:conf/icip/BewleyGORU16} and DSORT~\cite{DBLP:journals/corr/WojkeBP17}) and batch based methods (GOG~\cite{DBLP:conf/cvpr/PirsiavashRF11}, CEM~\cite{DBLP:journals/pami/MilanRS14}, SMOT~\cite{DBLP:conf/iccv/DicleCS13} and IOUT~\cite{DBLP:conf/avss/BochinskiES17}).

{\noindent \textbf{Metrics.}} We use multiple metrics to evaluate the MOT performance. These include identification precision (IDP)~\cite{DBLP:conf/eccv/RistaniSZCT16}, identification recall (IDR), and the corresponding F1 score IDF1 (the ratio of correctly identified detections over the average number of ground-truth and computed detections.),
Multiple Object Tracking Accuracy (MOTA)~\cite{DBLP:journals/ejivp/BernardinS08}, Multiple Object Tracking Precision (MOTP)~\cite{DBLP:journals/ejivp/BernardinS08}, Mostly Track targets (MT, percentage of groundtruth trajectories that are covered by a track hypothesis for at least $80\%$), Mostly Lost targets (ML, percentage of groundtruth objects whose trajectories are covered by the tracking output less than $20\%$), the total number of False Positives (FP), the total number of False Negatives (FN), the total number of ID Switches (IDS), and the total number of times a trajectory is Fragmented (FM).

{\noindent \textbf{Implementation Details.}} Since the above MOT algorithms are based on tracking-by-detection framework, all the $4$ detection inputs are provided for MOT task. We run them on all testing sequences of the \uav dataset on the machine with CPU i7 6700 and 32G memory, as well as a NVIDIA Titan X GPU.
\begin{table*}[t]
\centering
\scriptsize\setlength{\tabcolsep}{0.5pt}
\begin{tabular}{|c|ccc|cccc|cccc|c|}
\hline
MOT methods &IDF &IDP &IDR &MOTA &MOTP &MT[\%] &ML[\%] &FP &FN &IDS &FM &Speed [fps]\\
\hline
\multicolumn{13}{l}{Detection Input: Faster-RCNN~\cite{DBLP:conf/nips/RenHGS15}}\\
\hline
CEM~\cite{DBLP:journals/pami/MilanRS14} &$10.2 $ &$19.4  $ &$ 7.0$
&$-7.3$ &$69.6$ &$7.3$ &$68.6$ &$72,378$ &$290,962$ &$2,488$  &$\textbf{4,248}$ &$-/14.55$\\
CMOT~\cite{DBLP:conf/cvpr/BaeY14} &$ 52.0 $ &$63.9 $ &$43.8 $
&$36.4$ &$\textbf{74.5}$ &$36.5$ &$26.1$ &$ 53,920$ &$160,963$  &$1,777$ &$ 5,709$ &$-/2.83$\\
DSORT~\cite{DBLP:journals/corr/WojkeBP17} & $58.2 $ &$72.2$ &$ 48.8 $
&$40.7$ &$73.2$ &$41.7$ &$23.7$ &$44,868 $ &$155,290 $ &$ 2,061 $ &$ 6,432$ &$15.01/2.98$\\
\textbf{GOG}~\cite{DBLP:conf/cvpr/PirsiavashRF11}   & $0.4 $ &$ 0.5 $ &$ 0.3$
&$34.4$ &$72.2$ &$35.5$ &$25.3$ &$41,126$ &$168,194$ &$14,301$  &$12,516$ &$-/436.52$\\
\textbf{IOUT}~\cite{DBLP:conf/avss/BochinskiES17}  & $23.7 $& $30.3$& $ 19.5$
&$36.6$ &$72.1$ &$37.4$ &$25.0$ &$42,245$  &$163,881$ &$9,938$ &$10,463$ &$-/\textbf{1438.34}$\\
MDP~\cite{DBLP:conf/iccv/XiangAS15} &$ \textbf{61.5}$&$ \textbf{74.5}$&$ \textbf{52.3}$
&$\textbf{43.0}$ &$73.5$ &$\textbf{45.3}$ &$\textbf{22.7}$ &$ 46,151 $&$\textbf{147,735}  $&$ \textbf{541} $&$ 4,299$ &$-/0.68$\\
\textbf{SMOT}~\cite{DBLP:conf/iccv/DicleCS13}  &$ 45.0 $&$ 55.7$ &$ 37.8$
&$33.9$ &$72.2$ &$36.7$ &$25.7$ &$57,112$ &$166,528$ &$1,752$  &$9,577$ &$-/115.27$\\
\textbf{SORT}~\cite{DBLP:conf/icip/BewleyGORU16}  &$43.7$ &$58.9$ &$34.8$
&$39.0$ &$74.3$ &$33.9$ &$28.0$ &$\textbf{33,037}$  &$172,628$ &$2,350$ &$5,787$ &$-/245.79$\\
\hline

\hline
\multicolumn{13}{l}{Detection Input: R-FCN~\cite{DBLP:conf/nips/DaiLHS16}}\\
\hline

CEM~\cite{DBLP:journals/pami/MilanRS14}  &$10.3 $ &$18.4 $ &$ 7.2$
&$-9.6$ &$70.4$ &$6.0$ &$67.8$ &$81,617$ &$289,683$ &$2,201$  &$3,789$ &$-/9.82$\\
CMOT~\cite{DBLP:conf/cvpr/BaeY14} &$ 50.8 $&$59.4$ &$44.3$
&$ 27.1$ &$\textbf{78.5}$ &$35.9$ &$27.9$ &$ 80,592$ &$ 167,043 $ &$  919 $ &$ 2,788$ &$-/2.65$\\
DSORT~\cite{DBLP:journals/corr/WojkeBP17} &$55.5$ &$\textbf{67.3}$ &$47.2$
&$\textbf{30.9} $ &$ 77.0$ &$36.6$ &$27.4$ &$66,839$ &$ 168,409 $ &$  424  $ &$4,746$ &$9.22/1.95$\\
\textbf{GOG}~\cite{DBLP:conf/cvpr/PirsiavashRF11}  &$ 0.3 $ &$ 0.4  $ &$0.3$
&$28.5$ &$77.1$ &$34.4$ &$28.6$ &$60,511$ &$176,256$ &$6,935$  &$6,823$ &$-/433.94$\\
\textbf{IOUT}~\cite{DBLP:conf/avss/BochinskiES17}  &$44.0$ &$47.5$ &$40.9$
&$26.9$ &$75.9$ &$\textbf{44.3}$ &$\textbf{22.9}$ &$98,789$  &$\textbf{145,617}$ &$4,903$ &$6,129$ &$-/\textbf{863.53}$\\
MDP~\cite{DBLP:conf/iccv/XiangAS15}  &$\textbf{55.8}$ &$63.9$ &$\textbf{49.5}$
&$28.9 $ &$ 76.7$ &$40.9$ &$25.9$ &$82,540$ &$ 159,452$ &$   \textbf{411} $ &$ \textbf{2,705}$ &$-/0.67$\\
\textbf{SMOT}~\cite{DBLP:conf/iccv/DicleCS13}  &$44.0$ &$53.5$&$ 37.3$
&$24.5$ &$77.2$ &$33.7$ &$29.2$ &$76,544$ &$179,609$ &$1,370$  &$5,142$ &$-/64.68$\\
\textbf{SORT}~\cite{DBLP:conf/icip/BewleyGORU16} &$ 42.6$ &$58.7$ &$33.5$
&$30.2$ &$\textbf{78.5}$ &$29.5$ &$31.9$ &$\textbf{44,612}$  &$190,999$ &$2,248$ &$4,378$ &$-/209.31$\\
\hline

\hline
\multicolumn{13}{l}{Detection Input: SSD~\cite{DBLP:conf/eccv/LiuAESRFB16}}\\
\hline

CEM~\cite{DBLP:journals/pami/MilanRS14} &$ 10.1$ &$ 21.1$ &$  6.6$
&$-6.8$ &$70.4$ &$6.6$ &$74.4$ &$64,373$ &$298,090$ &$1,530$  &$\textbf{2,835}$ &$-/11.62$\\
CMOT~\cite{DBLP:conf/cvpr/BaeY14} &$ 49.4$ &$53.4 $&$46.0$
&$27.2 $ &$ 75.1$ &$38.3$ &$23.5$ &$98,915 $ &$146,418 $ &$ 2,920  $ &$6,914$ &$-/0.90$\\
DSORT~\cite{DBLP:journals/corr/WojkeBP17} &$51.4$ &$\textbf{65.7}$&$ 42.2$
&$33.6 $ &$ \textbf{76.7}$ &$27.9$ &$26.9$ &$\textbf{51,549 }$ &$173,639  $ &$\textbf{1,143} $ &$ 8,655$ &$15.00/3.46$\\
\textbf{GOG}~\cite{DBLP:conf/cvpr/PirsiavashRF11}  &$ 0.3 $ &$0.4$  &$0.3$
&$33.6$ &$76.4$ &$36.0$ &$22.4$ &$70,080$ &$148,369$ &$7,964$  &$10,023$ &$-/239.60$\\
\textbf{IOUT}~\cite{DBLP:conf/avss/BochinskiES17} &$29.4$ &$34.5$ &$25.6$
&$33.5$ &$76.6$ &$34.3$ &$23.4$ &$65,549$  &$154,042$ &$6,993$ &$8,793$ &$-/\textbf{976.47}$\\
MDP~\cite{DBLP:conf/iccv/XiangAS15} &$\textbf{58.8}$ &$63.2$ &$\textbf{55.0}$
&$\textbf{39.8} $ &$ 76.5$ &$\textbf{47.3}$ &$\textbf{19.5}$ &$ 79,760$ &$ \textbf{124,206 }$ &$ 1,310 $ &$ 4,539$ &$-/0.13$\\
SMOT~\cite{DBLP:conf/iccv/DicleCS13}  &$41.9$ &$45.9$ &$38.6$
&$27.2$ &$76.5$ &$34.9$ &$22.9$ &$95,737$ &$149,777$ &$2,738$  &$9,605$ &$-/11.59$\\
\textbf{SORT}~\cite{DBLP:conf/icip/BewleyGORU16}  &$37.1$ &$45.8$ &$31.1$
&$33.2$ &$\textbf{76.7}$ &$27.3$ &$25.4$ &$57,440$  &$166,493$ &$3,918$ &$7,898$ &$-/153.70$\\
\hline

\hline
\multicolumn{13}{l}{Detection Input: RON~\cite{DBLP:conf/cvpr/KongSYLLC17}}\\
\hline

CEM~\cite{DBLP:journals/pami/MilanRS14}  &$10.1$ &$18.8$  &$6.9$
&$-9.7$ &$68.8$ &$6.9$ &$72.6$ &$78,265 $ &$293,576  $ &$2,086  $ &$\textbf{3,526}$ &$-/9.98$\\
CMOT~\cite{DBLP:conf/cvpr/BaeY14}  &$57.5$ &$65.7$ &$51.1$
&$36.9 $ &$ \textbf{74.7}$ &$\textbf{46.5}$ &$\textbf{24.6}$ &$69,109 $ &$\textbf{144,760}$ &$1,111 $ &$ 3,656$ &$-/0.94$\\
DSORT~\cite{DBLP:journals/corr/WojkeBP17} &$58.3$ &$67.9$ &$51.2$
&$35.8$ &$71.5$ &$43.4$ &$25.7$ &$67,090 $  &$151,007   $  &$698  $  &$4,311$ &$17.45/4.02$\\
\textbf{GOG}~\cite{DBLP:conf/cvpr/PirsiavashRF11}  &$ 0.3$  &$0.3$ &$ 0.2$
&$35.7$ &$72.0$ &$43.9$ &$26.2$ &$62,929$ &$153,336$ &$3,104$  &$5,130$ &$-/287.97$\\
\textbf{IOUT}~\cite{DBLP:conf/avss/BochinskiES17}  &$50.1$ &$59.1$ &$43.4$
&$35.6$ &$72.0$ &$43.9$ &$26.2$ &$63,086$  &$153,348$ &$2,991$ &$5,103$ &$-/\textbf{1383.33}$\\
MDP~\cite{DBLP:conf/iccv/XiangAS15}  &$\textbf{59.9}$ &$\textbf{69.0}$ &$\textbf{52.9}$
&$35.3 $ &$ 71.7$ &$45.0$ &$25.5$ &$70,186$ &$ 149,980  $ &$ \textbf{414} $ &$ 3,640$ &$-/0.12$\\
SMOT~\cite{DBLP:conf/iccv/DicleCS13}  &$52.6$ &$60.8$ &$46.3$
&$32.8$ &$72.0$ &$43.4$ &$27.1$ &$73,226$ &$154,696$ &$1,157$  &$4,643$ &$-/29.37$\\
\textbf{SORT}~\cite{DBLP:conf/icip/BewleyGORU16}  &$54.6$ &$66.9$ &$46.1$
&$\textbf{37.2}$ &$72.2$ &$40.8$ &$28.0$ &$\textbf{53,435}$  &$159,347$ &$1,369$ &$3,661$ &$-/230.55$\\
\hline
\end{tabular}
\caption{Quantitative comparison results of MOT methods in the testing set of the \uav dataset. The last column shows the GPU/CPU speed. The best performer and realtime methods ($>30$fps) are highlighted in bold font. ``$-$'' indicates the data is not available.}
  \label{tab:mot_overall}
\end{table*}

\subsubsection{Overall Evaluation}
As shown in Table~\ref{tab:mot_overall}, MDP with Faster-RCNN has the best $43.0$ MOTA score and $61.5$ IDF score among all the combinations. Besides, the MOTA score of SORT in our dataset is much lower than other datasets with Faster-RCNN, \eg, $59.8\pm10.3$ in MOT16~\cite{DBLP:journals/corr/Anton16}. As object density is large in UAV videos, the FP and FN values on our dataset are also much larger than other datasets for the same algorithm. Meanwhile, IDS and FM appear more frequently. It means the proposed dataset is more challenging than existing ones.

Moreover, the algorithms using only position information (\eg, IOUT, SORT) could keep fewer tracklets combining with higher IDS and FM because of absence of appearance information. GOG has the worst IDF even though the MOTA is well because of the too much IDS and FM. DSORT performs well on IDS among these methods, which means deep feature has an advantage in the aspect of representing appearance of the same target. MDP mostly has the best IDS and FM value because of their individual-wised tracker model. So the trajectories are more complete than others with the higher IDF. Meanwhile, FP values will increase by associating more objects in complex scenes.
\begin{figure}[t]
\centering
\includegraphics[width=.95\linewidth]{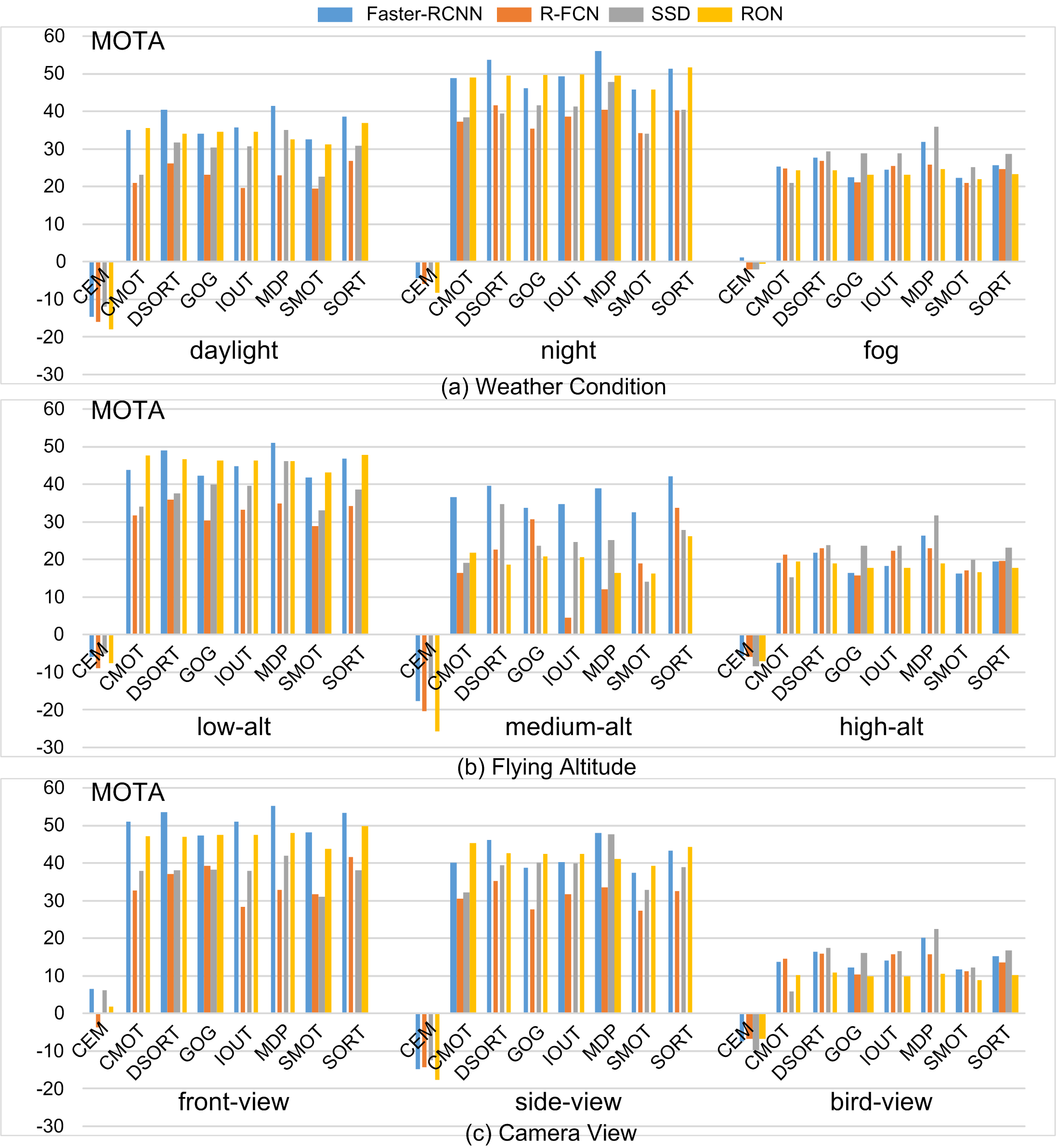}
  \caption{Quantitative comparison results of MOT methods in each attribute.}
\label{fig_attr_mot}
\end{figure}
\subsubsection{Attribute-based Evaluation}
Figure~\ref{fig_attr_mot} shows the performances of MOT methods on different attributes. Most methods perform better in \textit{daylight} than \textit{night} or \textit{fog} (see Figure~\ref{fig_attr_mot}(a)). It is fair and reasonable that objects in \textit{daylight} provide clearer appearance clues for tracking. In other illumination conditions, object appearance is confusing so the algorithms considering more motion clues achieve better performance, \eg, SORT, SMOT and GOG. Notably, on the sequences with \textit{night}, the performances of methods are much worse even the provided detections in \textit{night} own a good AP score. This is because objects are hard to track with confusing environment in \textit{night}. In Figure~\ref{fig_attr_mot}(b), the performance of most MOT methods increases with the decline of height. When UAVs capture videos in a lower height, fewer objects are captured in that view to facilitate object association. In terms of Camera Views as shown in Figure~\ref{fig_attr_mot}(c), vehicles in \textit{front-view} and \textit{side-view} offer more details to distinguish different targets compared with \textit{bird-view}, leading to better accuracy.

Besides, different detection input can guide MOT methods to focus on different scenes. Specifically, the performance with Faster-RCNN is better on sequences where object details are clearer (\eg, \textit{daylight}, \textit{low-alt} and \textit{side-view}); while R-FCN detection offers more stable inputs for each method when sequences have other challenging attributes, such as \textit{fog} and \textit{high-alt}. SSD and RON offer more accurate detection candidates for tracking such that the performances of MOT methods with these detections are balanced in each attribute.

{\noindent \textbf{Run-time Performance.}} Given different detection inputs, the speed of each method varies with the number of object detection candidates. However, IOUT and SORT using only position information generally proceed at ultra-real-time speed, while DSORT and CMOT using appearance information proceed much slower. As the object number is huge in our dataset, the speed of the method processing each object respectively (\eg, MDP) dramatically declines.
\begin{figure*}[t]
\centering
\includegraphics[width=.95\linewidth]{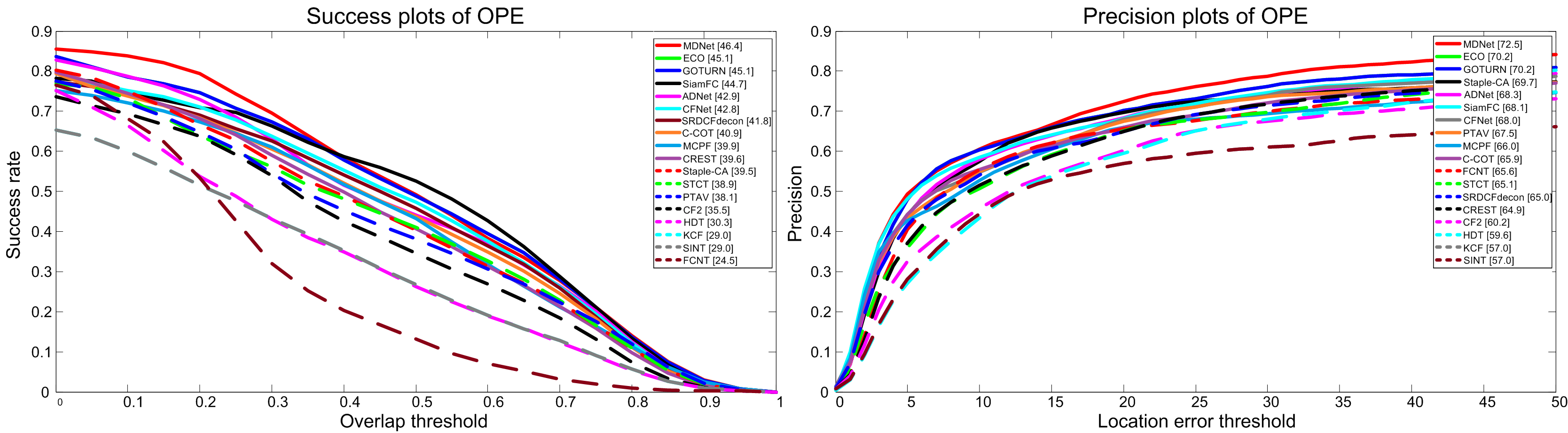}
  \caption{\small The precision and success plots on the \uav-SOT benchmark using One-pass Evaluation~\cite{DBLP:journals/pami/WuLY15}.}
\label{fig_sot_overall}
%~\vspace{-2em}
\end{figure*}
\begin{table*}[t]
\centering
\tiny\setlength{\tabcolsep}{.5pt}
\begin{tabular}{|c|cccccccc|c|}
  \hline
  SOT methods & \textbf{BC} & \textbf{CR} & \textbf{OR} & \textbf{SO} & \textbf{IV} & \textbf{OB} & \textbf{SV} & \textbf{LO} & Speed [fps]\\\hline
  MDNet~\cite{DBLP:conf/cvpr/NamH16}  & $\textbf{39.7}/\textbf{63.6}$ & $\textbf{43.0}/\textbf{69.6}$ & $\textbf{42.7}/\textbf{66.8}$ & $44.4/78.4$ & $\textbf{48.5}/76.4$ & $\textbf{47.0}/72.4$ & $\textbf{46.2}/\textbf{68.5}$ & $\textbf{38.1}/\textbf{54.7}$ &$0.89/0.28$\\
  ECO~\cite{DBLP:journals/corr/DanelljanBKF16} & $38.9/61.1$ & $42.2/64.4$ & $39.5/62.7$ & $\textbf{46.1}/79.1$ & $47.3/76.9$ & $43.7/71.0$ & $43.1/63.2$ & $36.0/50.8$  &$16.95/3.90$\\
  \textbf{GOTURN}~\cite{DBLP:conf/eccv/HeldTS16} & $38.9/61.1$ & $42.2/64.4$ & $39.5/62.7$ & $\textbf{46.1}/79.1$ & $47.3/76.9$ & $43.7/71.0$ & $43.7/63.2$ & $36.0/50.8$  &$\textbf{65.29}/11.70$\\
  \textbf{SiamFC}~\cite{DBLP:conf/eccv/BertinettoVHVT16} & $38.6/57.8$ & $40.9/61.6$ & $38.4/60.0$ & $43.9/73.2$ & $47.4/74.2$ & $45.3/\textbf{73.8}$ & $42.4/60.4$ & $35.9/47.9$  &$38.20/5.50$\\
  ADNet~\cite{DBLP:conf/cvpr/YunCYYC17} & $37.0/60.4$ & $39.9/64.8$ & $36.8/60.1$ & $43.2/77.9$ & $45.8/73.7$ & $42.8/68.9$ & $40.9/61.2$ & $35.8/49.2$   &$5.78/2.42$\\
  CFNet~\cite{DBLP:journals/corr/ValmadreBHVT17}& $36.0/56.7$ & $39.7/64.3$ & $36.9/59.9$ & $43.5/77.5$ & $45.1/72.7$ & $43.5/71.7$ & $40.9/61.1$ & $33.3/44.7$   &$8.94/6.45$\\
  SRDCF~\cite{DBLP:conf/iccv/DanelljanHKF15} & $35.3/58.2$ & $39.0/64.2$ & $36.5/60.0$ & $42.2/76.4$ & $45.1/74.7$ & $41.7/70.6$ & $40.2/59.6$ & $32.7/46.0$   &$-/14.25$\\
  SRDCFdecon~\cite{DBLP:conf/cvpr/DanelljanHKF16} & $36.0/57.4$ & $39.0/61.0$ & $36.6/57.8$ & $43.1/73.8$ & $45.5/72.3$ & $42.9/69.5$ & $38.0/54.9$ & $31.5/42.5$   &$-/7.26$\\
  C-COT~\cite{DBLP:conf/eccv/DanelljanRKF16} & $34.0/55.7$ & $39.0/62.3$ & $34.1/56.1$ & $44.2/79.2$ & $41.6/72.0$ & $37.2/66.2$ & $37.9/55.9$ & $33.5/46.0$   &$0.87/0.79$\\
  MCPF~\cite{DBLP:conf/cvpr/ZhangXY17} & $31.0/51.2$ & $36.3/59.2$ & $33.0/55.3$ & $39.7/74.5$ & $42.2/73.1$ & $42.0/73.0$ & $35.9/55.1$ & $30.1/42.5$   &$1.84/0.89$ \\
  CREST~\cite{DBLP:journals/corr/abs-1708-00225} & $33.6/56.2$ & $38.7/62.1$ & $35.4/55.8$ & $38.3/74.2$ & $40.5/69.0$ & $37.7/65.6$ & $36.5/56.7$ & $35.1/49.7$   &$2.83/0.36$\\
  \textbf{Staple-CA}~\cite{DBLP:conf/cvpr/MuellerSG17} & $32.9/59.2$ & $35.2/65.8$ & $34.6/62.0$ & $38.0/\textbf{79.6}$ & $43.1/\textbf{77.2}$ & $40.6/71.3$ & $36.7/62.3$ & $32.5/49.6$  &$-/\textbf{42.53}$\\
  STCT~\cite{DBLP:conf/cvpr/WangOWL16} & $33.3/56.0$ & $36.0/61.3$ & $34.3/57.5$ & $38.3/71.0$ & $40.8/69.9$ & $37.0/63.3$ & $37.3/59.9$ & $31.7/46.6$   &$1.76/0.09$ \\
  PTAV~\cite{DBLP:conf/iccv/HengH17} & $31.2/57.2$ & $35.2/63.9$ & $30.9/56.4$ & $38.0/79.1$ & $38.1/69.6$ & $36.7/66.2$ & $33.3/56.5$ & $32.9/50.3$   &$12.77/0.10$ \\
  CF2~\cite{DBLP:conf/iccv/MaHYY15} & $29.2/48.6$ & $34.1/56.9$ & $29.7/48.2$ & $35.6/69.5$ & $38.7/67.9$ & $35.8/65.1$ & $29.0/45.3$ & $28.3/38.1$   &$8.07/1.99$\\
  HDT~\cite{DBLP:conf/cvpr/QiZQYHL016} & $25.1/50.1$ & $27.3/56.2$ & $24.8/48.7$ & $29.8/72.6$ & $31.3/68.6$ & $30.3/65.4$ & $25.0/45.2$ & $25.4/37.6$   &$5.25/1.72$ \\
  \textbf{KCF}~\cite{DBLP:journals/pami/HenriquesC0B15} & $23.5/45.8$ & $26.7/53.4$ & $24.4/45.4$ & $25.1/58.1$ & $31.1/65.7$ & $29.7/65.2$ & $25.4/49.0$ & $22.8/34.4$  &$-/39.26$\\
  \textbf{SINT}~\cite{DBLP:conf/cvpr/TaoGS16} & $38.9/45.8$ & $26.7/53.4$ & $24.4/45.4$ & $25.1/58.1$ & $31.1/65.7$ & $29.7/65.2$ & $25.4/49.0$ & $22.8/34.4$   &$37.60/-$\\
  FCNT~\cite{DBLP:conf/iccv/WangOWL15} & $20.6/54.8$ & $21.8/60.2$ & $23.6/54.9$ & $21.9/71.9$ & $25.5/72.1$ & $24.2/70.5$ & $24.6/57.5$ & $22.3/47.2$   &$3.09/-$ \\
  \hline
\end{tabular}
\caption{Quantitative comparison results (\ie, overlap score$/$precision score) of SOT methods in each attribute. The last column shows the GPU/CPU speed. The best performer and realtime methods ($>30$fps) are highlighted in bold font. ``$-$'' indicates the data is not available.}
\label{tab:sot_attr}
\end{table*}
\subsection{Single Object Tracking}
The SOT field is dominated by correlation filter and deep learning based approaches~\cite{DBLP:conf/eccv/KristanLMFPCVHL16}. We evaluate $18$ recent such trackers on our dataset. These trackers can be generally categorized into $3$ classes based on their learning strategy and utilized features: I) correlation filter (CF) trackers with hand crafted features (KCF~\cite{DBLP:journals/pami/HenriquesC0B15}, Staple-CA~\cite{DBLP:conf/cvpr/MuellerSG17}, and SRDCFdecon~\cite{DBLP:conf/cvpr/DanelljanHKF16});
II) CF trackers with deep features (ECO~\cite{DBLP:journals/corr/DanelljanBKF16},  C-COT~\cite{DBLP:conf/eccv/DanelljanRKF16}, HDT~\cite{DBLP:conf/cvpr/QiZQYHL016}, CF2~\cite{DBLP:conf/iccv/MaHYY15}, CFNet~\cite{DBLP:journals/corr/ValmadreBHVT17}, and PTAV~\cite{DBLP:conf/iccv/HengH17});
III) Deep trackers (MDNet~\cite{DBLP:conf/cvpr/NamH16}, SiamFC~\cite{DBLP:conf/eccv/BertinettoVHVT16}, FCNT~\cite{DBLP:conf/iccv/WangOWL15}, SINT~\cite{DBLP:conf/cvpr/TaoGS16}, MCPF~\cite{DBLP:conf/cvpr/ZhangXY17}, GOTURN~\cite{DBLP:conf/eccv/HeldTS16}, ADNet~\cite{DBLP:conf/cvpr/YunCYYC17}, CREST~\cite{DBLP:journals/corr/abs-1708-00225}, and STCT~\cite{DBLP:conf/cvpr/WangOWL16}).

{\noindent \textbf{Metrics.}} Following the popular visual tracking benchmark~\cite{DBLP:journals/pami/WuLY15}, we adopt the success plot and precision plot to evaluate the tracking performance. The success plot shows the percentage of bounding boxes whose intersection over union with their corresponding groundtruth bounding boxes are larger than a given threshold. The trackers in success plot are ranked according to their \textit{success score}, which is defined as
the area under the curve (AUC). The precision plot presents the percentage of bounding boxes whose center points are within a given distance ($0\sim50$ pixels) to the ground truth. Trackers in precision plot are ranked according to their \textit{precision score}, which is the percentage of bounding boxes within a distance threshold of $20$ pixels.

{\noindent \textbf{Implementation Details.}} All the trackers are run on the machine with CPU i7 4790k and 16G memory, as well as a NVIDIA Titan X GPU.

\subsubsection{Overall Evaluation}
The performance for each tracker is reported in Figure~\ref{fig_sot_overall}. The figure shows that: I) All the evaluated trackers perform not well on our dataset. Specifically, the state-of-the-art methods such as MDNet only achieves $46.4$ success score and $72.5$ precision score. Compared to the best results (\ie, $69.4$ success score and $92.8$ precision score) on OTB100~\cite{DBLP:journals/pami/WuLY15}, a significantly large performance gap is formulated. Such performance gap is also observed when compared to the results on UAV-123. For example, KCF achieves a success score of $33.1$ on UAV-123 but only $29.0$ on our dataset. These results indicate that our dataset poses new challenges for the visual tracking community and more efforts can be devoted to the real-world UAV tracking task. II) Generally, deep trackers achieves more accurate results than CF trackers with deep features, and then CF trackers with hand-crafted features. Among the top $10$ trackers, there are $6$ deep trackers (MDNet, GOTURN, SianFC, ADNet, MCFP and CREST), $3$ CF trackers with deep features (ECO, CFNet, and C-COT), and one CF tracker with hand-crafted features namely SRDCFdecon.
\subsubsection{Attribute-based Evaluation}
As presented in Table~\ref{tab:sot_attr}, the deep tracker MDNet achieves best results on $7$ out of $8$ tracking attributes, which can be attributed to its multiple domain training and hard sample mining. CF trackers with deep features such as CF2 and HDT fall behind due to no scale adaptation. SINT~\cite{DBLP:conf/cvpr/TaoGS16} does not update its models during tracking, which results in a limited performance. Staple-CA performs well on the \textbf{SO} and \textbf{IV} attributes, as its improved model update strategy can reduce over-fitting to recent samples. Most of the evaluated methods act poorly on the \textbf{BC} and \textbf{LO} attributes, which may be caused by the decline of discriminative ability of appearance features extracted from cluttered or low resolution image regions.

{\noindent \textbf{Run-time Performance.}} From the last column of Table~\ref{tab:sot_attr}, We note that I) The top $10$ accurate trackers run far from real time even on a high-end CPU. For example, the fastest tracker among top $10$ accurate only runs at $11.7$fps and the most accurate MDNet runs at $0.28$ fps. On the other hand, the realtime trackers on CPU (\eg, Staple-CA and KCF), achieve success scores $39.5$ and $29.0$, which are intolerant for practical applications. II) When a high-end GPU card is used, only $3$ out of $18$ trackers (GOTURN, SiamFC, SINT) can perform in real-time. But again their best success score is just $45.1$, which is not accurate enough for real applications. Overall, more work need to be done to develop a faster and more precise tracker.
%------------------------------------------------------------------------
\section{Discussion}
Our benchmark, delivering from real-life demand, vividly samples real circumstances. Since algorithms generally perform poorly on it comparing with their plausible performances with other datasets, we think this benchmark dataset can reveal some promising research trends and benefit the community. Based on the above analysis, there are several research directions worth exploring:
\begin{itemize}
  \item \textbf{Realtime issues.} Running speed is a crucial measurement in practical applications. Although the performance of deep learning approaches surpass other methods by a large margin (especially in SOT task), the requirements of computational resources are very harsh in embedded UAV platforms. To achieve high efficiency while maintaining comparable accuracy, some recent methods~\cite{DBLP:conf/cvpr/ZhangZMHS15,DBLP:conf/nips/WenWWCL16} develop an approximate network by pruning, compressing, or low-bit representing. We expect the future works count more real-time constraints not just accuracy.
  \item \textbf{Scene priors.} Different methods perform the best in different scenarios. When considering scene priors in detection and tracking approaches, more robust performance is expected. For example, MDNet~\cite{DBLP:conf/cvpr/NamH16} trains a specific object-background classifier for each sequence to handle varies scenarios, which make it rank the first in most datasets. We think along with our dataset this magnificent design may inspired more methods to deal with mutable scenes.
  \item \textbf{Motion clues.} Since the appearance information is not always reliable, the methods evaluated in our dataset would gain more robustness when considering motion clues. Many recently proposed algorithms make their efforts in this trend with the help of LSTM~\cite{DBLP:conf/iccvw/YangC17,DBLP:conf/cvpr/KahouMMPV17}, but still have not met with expectations. Considering with the fierce motions of both object and background, our benchmark may fruit this research trend in the future.
  \item \textbf{Small objects.} In our dataset, $27.5\%$ of objects consist of less than $400$ pixels, almost $0.07\%$ of a frame. It provides limited textures and contours for feature extraction which causes the accuracy loss of algorithms heavily based on appearance. Meanwhile, generally methods tend to save their time consuming by down-sampling images. It exacerbates the situations harshly, \eg, DET methods mentioned above generally enjoy a $10\%$ accuracy rise due to our parameters adjusting of authors provided codes and settings, mainly dealing with the size of anchors. However their performance still cannot met with expectation. We advise researchers should gain more promotions if they pay more attention on handling with small objects.
\end{itemize}
\section{Conclusion}
In this paper, we construct a new and challenging UAV benchmark for $3$ foundational visual tasks including DET, MOT and SOT. The dataset consists of $100$ videos ($80k$ frames) captured with UAV platform from complex scenarios. All frames are annotated with manually labelled bounding boxes and $3$ circumstances attributes, \ie, weather condition, flying altitude, and camera view. SOT dataset has additional $8$ attributes, \eg, background clutter, camera rotation and small object. Moreover, an extensive evaluation of most recent and state-of-the-art methods is provided. We hope the proposed benchmark will contribute to the computer vision community by establishing a unified platform for evaluation of detection and tracking methods for real scenarios. In the future, we expect to extend the current dataset to include more sequences for other high-level tasks applied in computer vision, and richer annotations for evaluation on corresponding algorithms.

%\clearpage

\bibliographystyle{splncs03}
\bibliography{arxiv2018-uav-benchmark}
\end{document}